\documentclass{article}
\usepackage{spconf,amsmath,epsfig}
\usepackage[caption=false]{subfig}
\usepackage{multirow}
\usepackage{dsfont}
\usepackage{stfloats}
\usepackage{subfig} 
\usepackage{tabu}  
\usepackage{booktabs}
\usepackage{algorithm}
\usepackage{algorithmic}
\usepackage{titlesec}
\usepackage{mfirstuc}
\usepackage{titlecaps}
\usepackage{rotating}
\usepackage[table]{xcolor}
\usepackage{amsmath} 
\usepackage{cite}
\usepackage[pagebackref=false, breaklinks=true, colorlinks, urlcolor=gray,
            linkcolor=linkcolor, bookmarks=false]{hyperref}

\definecolor{dg}{RGB}{0,100,0}
\definecolor{bareland}{RGB}{128,0,0}
\definecolor{rangeland}{RGB}{0,255,36}
\definecolor{developed}{RGB}{148,148,148}
\definecolor{tree}{RGB}{34,97,38}
\definecolor{water}{RGB}{0,69,255}
\definecolor{agri}{RGB}{75,181,73}
\definecolor{building}{RGB}{222,31,7}
\definecolor{linkcolor}{HTML}{ED1C24}

\title{\titlecap{\textit{ChangeAnywhere}: Sample Generation for Remote Sensing Change Detection via Semantic Latent Diffusion Model}}
\name
 {Kai Tang, Jin Chen\sthanks{Corresponding Author (chenjin@bnu.edu.cn)
 }
}
\address{State Key Laboratory of Remote Sensing Science, Beijing Normal Unieversity}

\begin{document}
%
\maketitle
%
\noindent \textit{Remote sensing change detection (CD) is a pivotal technique that pinpoints changes on a global scale based on multi-temporal images. With the recent expansion of deep learning, supervised deep learning-based CD models have shown satisfactory performance. However, CD sample labeling is very time-consuming as it is densely labeled and requires expert knowledge. To alleviate this problem, we introduce ChangeAnywhere, a novel CD sample generation method using the semantic latent diffusion model and single-temporal images. Specifically, ChangeAnywhere leverages the relative ease of acquiring large single-temporal semantic datasets to generate large-scale, diverse, and semantically annotated bi-temporal CD datasets. ChangeAnywhere captures the two essentials of CD samples, i.e., change implies semantically different, and non-change implies reasonable change under the same semantic constraints. We generated \texttt{ChangeAnywhere-100K}, the largest synthesis CD dataset with 100,000 pairs of CD samples based on the proposed method. The \texttt{ChangeAnywhere-100K} significantly improved both zero-shot and few-shot performance on two CD benchmark datasets for various deep learning-based CD models, as demonstrated by transfer experiments. This paper delineates the enormous potential of ChangeAnywhere for CD sample generation, and demonstrates the subsequent enhancement of model performance. Therefore, ChangeAnywhere offers a potent tool for remote sensing CD. All codes and pre-trained models will be available at \small \hypersetup{urlcolor=magenta}\url{https://github.com/tangkai-RS/ChangeAnywhere}.}
%
\par
\section{Introduction}\label{sec:intro}
\vspace{-7pt}
\setlength{\parskip}{2.5pt} 

\par The Earth has been undergoing continuous anthropogenic and natural change. Spaceborne remote sensing change detection (CD) provides a unique opportunity to accurately reveal these changes on a planetary scale, by comparing multi-temporal images at the same geo-registered location but at different times \cite{KENNEDY20091382}. Inspired by the rapid advances of deep learning, there are various excellent deep learning-based CD models \cite{Zhu2017a}. The outstanding performance of the CD models is undoubtedly attributed to the vast volume of training data. However, it's worth noting that acquiring CD samples is an immensely laborious and time-consuming process.
\par To alleviate sample starvation in CD, numerous studies have employed synthetic samples. Three primary categories exist for generating synthetic CD samples: graphics-based, generative model-based, and data augmentation-based methods. The graphics-based methods simulate the remote sensing images by synthesizing a virtual world through 3D modeling software and obtaining a top view \cite{Song_2024_WACV, FUENTESREYES202374}. There is a domain gap between the samples generated in the graphics-based way and the real remote sensing images, which brings a great obstacle to the practical application of the samples. The generative model-based methods mainly utilize generative adversarial neural networks (GANs) to generate images or instances \cite{zheng2023changen}. However, GANs are prone to the problem of model collapse and unstable training due to the lack of theoretical understanding and thus do not always synthesize high-quality samples. The data augmentation-based methods construct bi-temporal CD samples by operations such as pairing or paste-copying single-temporal images \cite{seo2023selfpair, Zheng2021Change, CHEN202387}, which are prone to produce overly complex samples with negative classes that differ too much from real image pairs and are not conducive to the model's ability to learn CD generalized knowledge. In summary, the authenticity and quality of the samples produced by the three methods described above could be further improved. Additionally, existing CD sample synthesis methods typically synthesize change samples for only a single target of interest, such as building \cite{seo2023selfpair, Zheng2021Change, zheng2023changen}, and thus lack diversity. However, other types of change information such as forests and croplands, hold significant value for ecological and food security applications. Therefore, the application scope of existing methods cannot meet the vast demand.
\par Denoising diffusion probabilistic models (DDPMs) \cite{DBLP:journals/corr/abs-2006-11239} are an emerging generative model based on maximum likelihood learning that has shown promising results in Artificial Intelligence Generative Content (AIGC) \cite{cao2023comprehensive}, e.g., Stable Diffusion \cite{rombach2021highresolution}, DALL-E \cite{DBLP:journals/corr/abs-2102-12092}, and Imagen Video \cite{ho2022imagen}. There are two processes in DDPMs, the forward diffusion process, which gradually transforms data into noise through a Markov chain, and the reverse denoising process, which allows for more controlled and stable model learning. Compared to the GANs, DDPMs have a well-established foundation in stochastic differential equations, providing a clearer theoretical understanding and could produce high-quality samples. More importantly, DDPMs are well suited for condition generation tasks such as text-to-image synthesis \cite{li2023snapfusion, zhang2023texttoimage} or super-resolution \cite{dong2024building, LI202247}. The diffusion process can naturally incorporate conditional information, resulting in controlled sample synthesis. Unfortunately, the potential of DDPMs for CD sample synthesis has not been explored.
\par To fill the above two research gaps (i.e., quality and diversity of CD samples), we propose a sample generation pipeline for remote sensing CD based on the DDPM and single temporal imagery, called ChangeAnywhere. Change Anything is inspired by the current state of remote sensing datasets i.e., it is easier to acquire large volume single temporal remote sensing semantic segmentation datasets than multi-temporal CD datasets. Meanwhile, our generation process captures the two essentials of CD samples, i.e., change implies semantically different, and non-change implies reasonable change under the same semantic constraints. To verify the validity of ChangeAnywhere, we generated a large CD dataset based on Change Anything and validated the zero-shot and few-shot performance after its pre-training on various CD models. The main contributions of this paper are summarized as follows:
\begin{itemize}
\item  We propose ChangeAnywhere, the first DDPM-based sample synthetic method for CD.
\item We generated \texttt{ChangeAnywhere-100K}, the largest synthetic dataset for the CD with bi-temporal semantic annotation, based on ChangeAnywhere.
\item The zero-shot and few-shot performance of the eight CD models on two real CD benchmark datasets is significantly improved after \texttt{ChangeAnywhere-100K} pre-training.	
\end{itemize}

\section{Methodology}\label{sec:method}
\vspace{-7pt}
\par
\begin{figure*}[!t]
  \centering
  \resizebox{0.9\linewidth}{!}{
  \includegraphics[width=\linewidth]{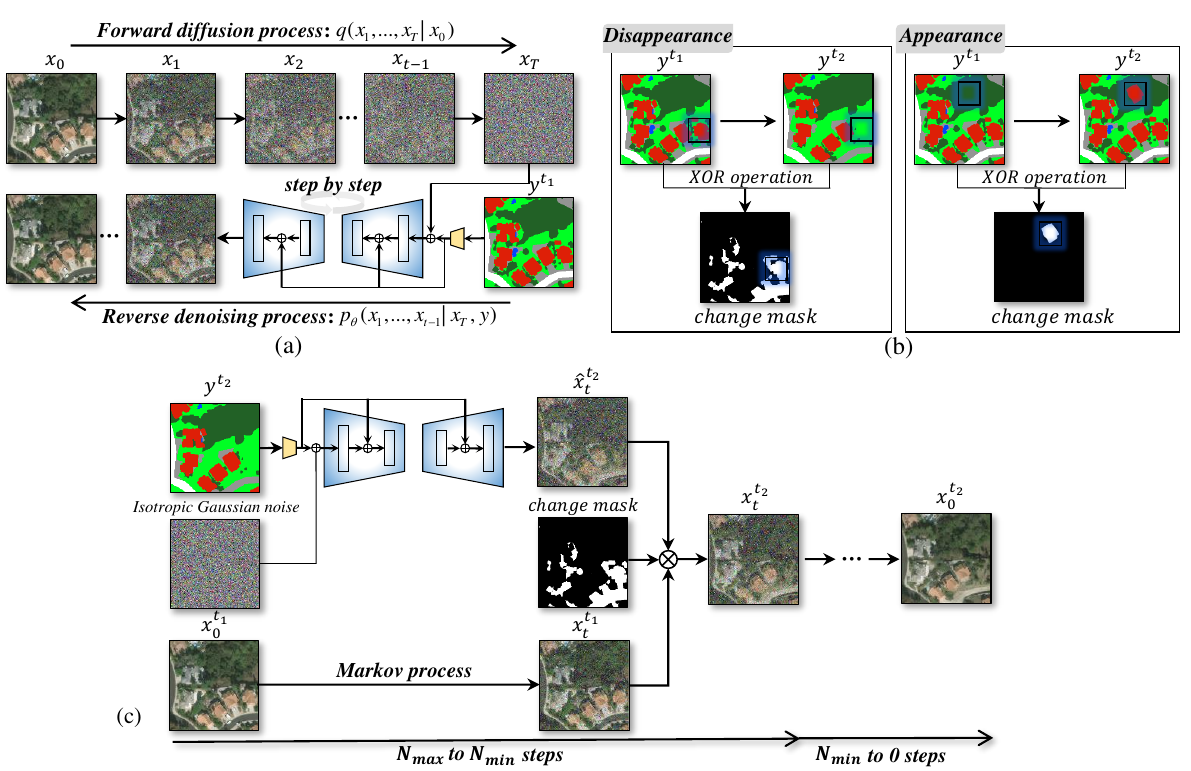}
  }
  \caption{
    The overall flow of ChangeAnywhere. (a) Semantic latent diffusion model training. (b) Change events simulation. (c) Change Sample Generation. All the diffusion and denoising processes are performed in the latent space, and we omit the process of encoding and decoding the original remote sensing images using VQGAN.
    }
    \label{fig-overall}
    \vspace{-14pt}
\end{figure*}

 The overall flow of ChangeAnywhere is shown in Fig. \ref{fig-overall}. We first trained a semantic control-based latent diffusion model (semantic latent diffusion model). The well-trained semantic latent diffusion model can generate high-quality remote sensing images based on semantic mask guidance. Second, we simulated the change events by changing the semantic mask ($y^{t_{1}}$) of the original image ($x^{t_{1}}$), and generating a changed semantic mask ($y^{t_{2}}$) and change mask. Finally, we input the $x^{t_{1}}$, $y^{t_{2}}$, and change mask into the semantic latent diffusion model to generate the image ($x^{t_{2}}$) that matches pre-defined change event.
\subsection{Preliminaries}
\vspace{-5pt}
\par DDPMs encompass both forward diffusion and reverse denoising processes \cite{DBLP:journals/corr/abs-2006-11239}. When presented with an image or latent variable $x_0$. Gaussian noise is incrementally introduced to $x_0$. This procedure adheres to a Markov process governed by a variance schedule $\beta_1$,...,$\beta_T$:
\begin{align}
  \label{eqn:1}
      \begin{split}
           q(x_1, ..., x_T | x_0) &= \prod_{t=1}^{T} q(x_t | x_{t-1})
          \\
           q(x_t | x_{t-1}) &= \mathcal{N}(\sqrt{1-\beta_t} x_{t-1}, \beta_t \mathbf{I})
      \end{split}
\end{align}
\noindent where $T$ is the total number of steps. As long as $T$ is large enough, it will be an isotropic Gaussian distribution. According to the Markov process, the $x_t$ at any step t in a closed form using the reparameterization trick:
\begin{align}
  \label{eqn:2}
      \begin{split}
           q(x_t|x_0) &= \mathcal{N}(\sqrt{\bar{\alpha}_t} x_0, (1-\bar{\alpha}_t) \mathbf{I})
          \\
           x_t &=  \sqrt{\bar{\alpha}_t} x_0 + \sqrt{1-\bar{\alpha}_t} \epsilon,
      \end{split}
\end{align}
\noindent where $\epsilon \sim \mathcal{N}(0,\mathbf{I})$, $\alpha_t = 1 - \beta_t$ and $\bar{\alpha}_t = \prod_{s=0}^{t} \alpha_s$.
\par To generate a clear and realistic image (i.e., to gradually strip the noise from an isotropic Gaussian noise) one needs to estimate $q(x_0|x_t)$. However, we cannot easily estimate it 
because the entire dataset needs to be used. 
But under $x_0$, and combining Bayes' theorem $q(x_{t-1)}|x_{t}, x_0) = q(x_t|x_{t-1}, x_0)\frac{q(x_{t-1)}|x_0)}{q(x_{t)}| x_0)}$, we can estimate the $\mu(x_t, t)$ (which can represent the average state of the real image):
\begin{align}
  \label{eqn:3}
      \begin{split}
        \mu(x_t, t) = \frac{1}{\sqrt{\alpha_t}} \left( x_t - \frac{\beta_t}{\sqrt{1-\bar{\alpha}_t}} \epsilon(x_t, t) \right)
      \end{split}
\end{align}
\par Then, we can then use a neural network $f(\theta)$ to approximate $\mu_(x_t, t)$, i.e., $p_{\theta}(x_0|x_t)$. Thus, the loss function aims to minimize $\mu_{\theta}(x_t, t)$ and $\mu_(x_t, t)$, which can be obtained through the simplification proposed by Ho et al. \cite{DBLP:journals/corr/abs-2006-11239}:
\begin{align}
  L = E_{t,x_0,\epsilon_t}[ || \epsilon_t - \epsilon_{\theta}(x_t, t) ||^2 ]
  \label{loss}
\end{align}
\par In the stepwise generation process $x_{t-1}$ can be computed as follows, i.e., 
taking into account both the average state of the real image and a suitable degree of randomness.
\begin{align}
  x_{t-1} = \frac{1}{\sqrt{\alpha_t}}\left(x_t - \frac{1-\alpha_t}{\sqrt{1-\bar\alpha_t}} \epsilon_\theta(x_t, t) \right) + {\sqrt{1-\bar{\alpha}_t}}z
\end{align}

\subsection{Semantic Latent Diffusion Model}
\vspace{-5pt}


\par 
For conditional DDPMs, such as text-to-image \cite{li2023snapfusion, zhang2023texttoimage}  or image-to-image models \cite{Saharia2021PaletteID}, conditions ($y$) can be integrated into the network through operations like concatenation or attention mechanisms, without changing the original loss function. The learning objective of the conditional DDPMs is $p_{\theta}(x_0|x_t, y)$, but the loss function is consistent with the original DDPMs (Eq. \ref{loss}). In the generation process, the model will generate clear images within the constraints of the conditional information.
\par In this paper, we utilized OpenEarthMap \cite{xia2022openearthmap}, a benchmark dataset designed for global high-resolution land cover mapping. It comprises 5000 aerial and satellite annotated masks with 8-class land cover labels (bare land, rangeland, developed space, road, tree, water, agriculture land, building), along with 2.2 million segments at a spatial resolution of 0.25-0.5 meters. The dataset spans 97 regions from 44 countries across 6 continents.
\par We trained a semantic diffusion model based on the latent diffusion model \cite{rombach2022high} and OpenEarthMap \cite{xia2022openearthmap}, where semantic masks ($y^{t_{1}}$) were used as conditions. Note that we trained a VQGAN \cite{yu2022vectorquantized} to encode the original remote sensing images spatially downsampled by a factor of 4 into the latent space and then trained the latent diffusion model. This can speed up the model training and sample generation \cite{rombach2021highresolution}. 

\subsection{Change Events Simulation}
\vspace{-5pt}
\par Inspired by two change events in the real world, i.e., the appearance and disappearance of targets at a time point, we designed a method that can automatically generate change masks. Specifically, leveraging the semantic mask $y^{t_{1}}$, for target appearance, the mask of a smaller instance within the semantic mask is randomly moved to the interior of the more homogeneous region. For target disappearance, the semantics of the selected instance is randomly changed to the semantics of its surrounding land cover. In both ways, we simulated numerous semantic masks $y^{t_{2}}$, and the change mask can be obtained by a simple XOR operation.
\subsection{Change Sample Generation}
\vspace{-5pt}
\par Assuming that the entire sample generation using semantic latent model is in $N$ steps and distributed in two phases $N_{max}$ to $N_{min}$ ($N_{min}>0$) and $N_{min}$ to 0. At the $N_{max}$ in the $N_{min}$ step, the existing image is $x_{0}^{t_{1}}$, and our goal is to generate $x_{0}^{t_{ 2}}$ such that the two images can form a pair of CD samples. Assume that the $x_{0}^{t_{2}}$ image has both change and non-change regions. To control the non-change region to keep the original semantics, we can easily compute $x_{t}^{t_{1}}$ based on the existing $x_{0}^{t_{1}}$ image and Markov process. Also, we can infer a $\hat{x}_{t}^{t_{2}}$ based on the well-trained semantic latent diffusion model. Then we perform a linear mixing of the two by the change mask ($Mask$) to get.
\begin{align}
    x_{t}^{t_{2}} = \hat{x}_{t}^{t_{2}} \odot Mask + {x}_{t}^{t_{1}} \odot (1-m)
\end{align}
\par Non-change regions of $t_2$ have features that are highly consistent with the features of the original image, while the change regions are generated semantically and constrained randomly under pre-defined change semantic constraints. At $N_{min}$ to step 0, $x_{t}^{t_{2}}$ will be generated completely randomly under the semantic constraints of $x_{0}^{t_{2}}$ to simulate plausible variations in the non-change regions, i.e., differences in phenology and differences in the imaging conditions of the satellite sensor. It's worth noting that all the above processes are carried out in latent space, and the final images are obtained by well-trained VQGAN decoding. Additionally, the strategy of Song et al. \cite{song2022denoising} was used in the generation process to speed up the efficiency of the generation.
\vspace{-7pt}

\section{Experiments}\label{sec:experiments}
\vspace{-7pt}

\subsection{Experimental Setting}
\vspace{-5pt}

\par \noindent \textbf{Semantic latent diffusion traning.} The VQGAN and semantic latent diffusion model were trained based on Pytorch and an NVIDIA A800 GPU. The VQGAN was trained with 100k iterations and the learning rate was set to 4.5e-6. The semantic latent diffusion model was trained with 500k iterations and the learning rate was set to 1e-6. The patch size was set to $256 \times 256$ and the batch size was set to 16.
\par \noindent \textbf{CD dataset generation.} We based on a total of 50,000 images and annotated masks from OpenEarthMap's train and part of the val subset and simulated the change events of target appearance and disappearance for each image. Then, we generated \texttt{ChangeAnywhere-100K} (hereafter referred to as \texttt{Ce-100K}) with 100,000 pairs of CD samples based on the proposed method.
\par \noindent \textbf{CD models traning.} Eight CD models (FC-EF \cite{CayeDaudt2018}, FC-Siam-conc \cite{CayeDaudt2018}, FC-Siam-diff \cite{CayeDaudt2018}, SNUNet \cite{Fang2022SNUNet}, BiT \cite{Chen2022Remote}, Changer \cite{fang2022changer}, ChangeFormer \cite{bandara2022transformerbased}, and LightCDNet \cite{10214556}) and two benchmark CD datasets (SECOND \cite{Yang2022Asymmetric} and SYSU-CD \cite{Shi2022Deeply}) were selected to validate the transfer (i.e., zero-shot and few-shot) performance after pre-training on \texttt{Ce-100K}). All models were trained on an NVIDIA GTX 3090 GPU and based on the Open-CD \cite{opencd2022}. During the pre-training, the initial learning rate was set to 1e-3, and the poly schedule was adopted. The cross-entropy loss and AdamW optimizer were used to optimize the parameters of the models. The weight decay of the AdamW optimizer was set to 0.05. The training image size is $256 \times 256$, and the batch size was set to 8. In the fine-tuning stage, we set the learning rate of all models to 1e-4 and trained 50 epochs. The other settings are consistent with the pre-training phase.
\par \noindent \textbf{Metrics.} F$_1$, Precision, Recall, and Intersection over Union (IoU) in terms of change class were used to evaluate the CD performance of the models.

\begin{figure}[t]
  \includegraphics[width=3.32in]{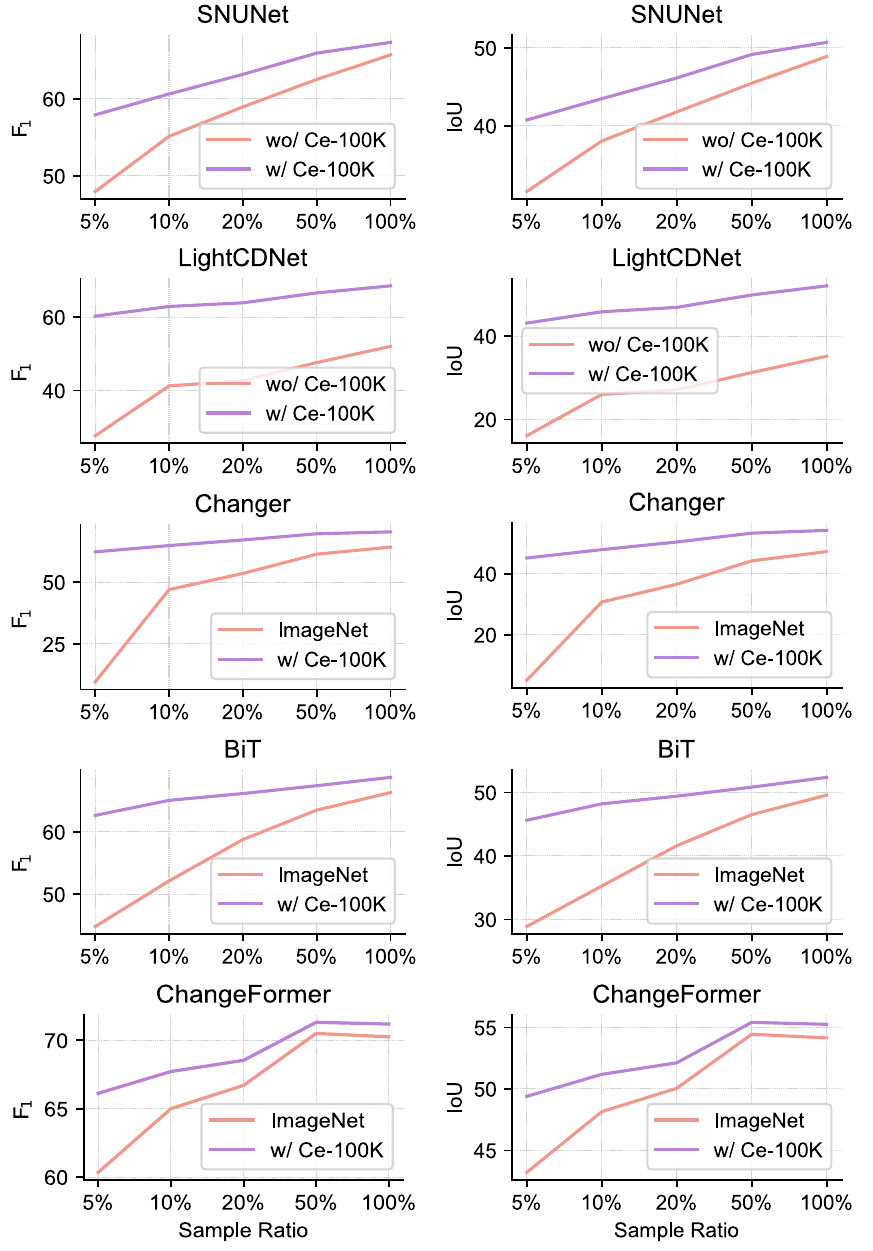}
  \caption{Various change detect models performance curves on the SECOND dataset.}
  \label{fig-line-second}
\end{figure}

\subsection{Visualization for \texttt{Ce-100K}}\label{sec:exper-Ce-100K}
\vspace{-5pt}
\par
\begin{figure*}[!t]
  \centering
  \resizebox{0.9\linewidth}{!}{
  \includegraphics[width=\linewidth]{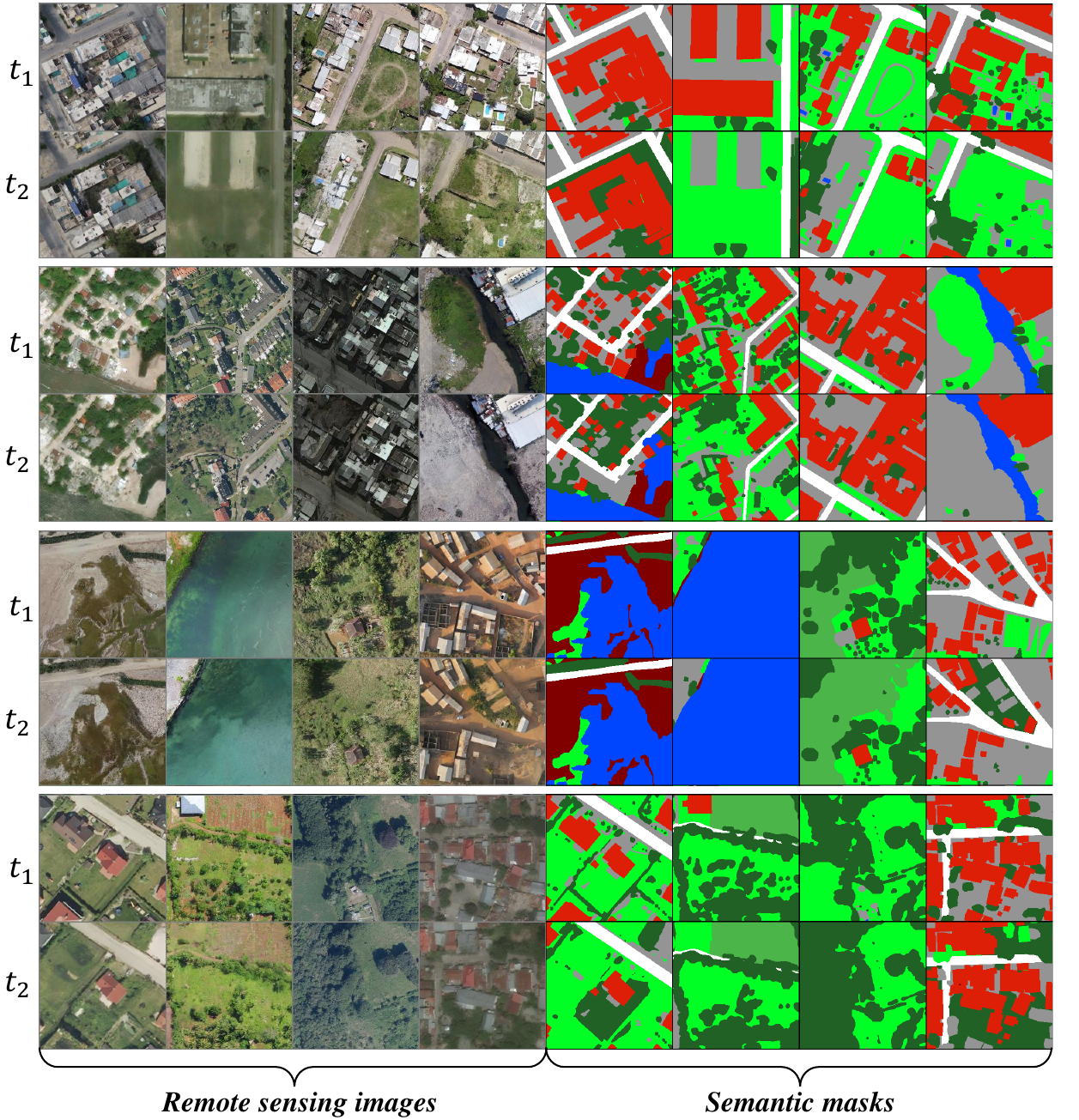}
  }
  \caption{
    Visualization of \texttt{ChangeAnywhere-100K}. Legend: \textcolor{tree}{\textbf{Tree}}, \textcolor{bareland}{\textbf{Bare land}}, \textcolor{rangeland}{\textbf{Rangeland}}, \textcolor{developed}{\textbf{Developed space}}, \textcolor{water}{\textbf{Water}}, \textcolor{agri}{\textbf{Agriculture land}}, \textcolor{building}{\textbf{Building}}, and white replaces Road.
    }
    \label{fig-visual}
    \vspace{-14pt}
\end{figure*}

 As shown in Fig. \ref{fig-visual}, \texttt{Ce-100K} includes very realistic $x^{t_{2}}$ in all kinds of change scenes and types. There are still reasonable variations in the non-change regions, such as shadows, phenological differences, and imaging angle differences. All these prove ChangeAnywhere's excellent ability to generate real CD datasets.

\begin{table*}[!ht]
  \setlength\tabcolsep{5pt} \small
  \renewcommand{\arraystretch}{1.2}
  \caption{\centering{Zero-shot performance evaluation of different methods on the two benchmark CD datasets. The highest values are marked in \textbf{bold} and the next highest values are \underline{underlined}.}}
  \centering
  \begin{tabular}{l |c| c c c c |c c c c}
    \hline
    \multirow{2}{*}{\textbf{Method}} & \multirow{2}{*}{\textbf{Backbone}} & \multicolumn{4}{c}{\textbf{SECOND}} & \multicolumn{4}{c}{\textbf{SYSU-CD}} \\
    \cline{3-10} 
    ~ & ~ & F$_1$ & Precision & Recall & IoU & F$_1$ & Precision & Recall & IoU \\  \hline\hline
    \textbf{\textit{ConvNet-based}} & ~ & ~ & ~ & ~ & ~ & ~ & ~ & ~ & ~ \\ 
    FC-EF \cite{CayeDaudt2018} & - & 36.37 & 22.48 & 95.17 & 22.23 & 52.27 & 36.93 & \textbf{89.41} & 35.38 \\ 
    FC-Siam-Conc \cite{CayeDaudt2018} & - & 35.91 & 22.48 & 89.22 & 21.89 & 50.45 & 36.31 & 82.67 & 33.74 \\ 
    FC-Siam-Diff \cite{CayeDaudt2018} & - & 37.47 & 25.04 & 74.38 & 23.05 & 53.46 & 44.23 & 67.54 & 36.48 \\
    SNUNet \cite{Fang2022SNUNet} & - & 36.07 & 22.17 & \textbf{96.59} & 22.00 & 53.93 & 38.95  & 87.63  & 36.92  \\ 
    LightCDNet \cite{10214556})  & - & 35.88 & 22.06 & \underline{96.08} & 21.86 & \underline{55.07} & 40.09  & \underline{87.92}  & \underline{38.00}  \\ 
    Changer \cite{fang2022changer} & ResNet18 & 38.17 & 23.98 & 93.48 & 23.58 & \textbf{58.65} & \underline{44.86} & 84.69  & \textbf{41.49}  \\ 
    \textbf{\textit{Transformer-based}} & ~ & ~ & ~ & ~ & ~ & ~ & ~ & ~ & ~ \\ 
    BiT \cite{Chen2022Remote} & ResNet18 & \textbf{47.54} & \textbf{37.92} & 63.73  & \textbf{31.19}  & 53.15  & \textbf{52.10}  & 54.25  & 36.19  \\ \
    ChangeFormer \cite{bandara2022transformerbased} & MiT-B1 & \underline{40.79} & \underline{26.67} & 86.66  & \underline{25.62}  & 52.83  & 38.76  & 82.94  & 35.90  \\   \hline
  \end{tabular}
  \label{table:zero-shot}
\end{table*}

\par \noindent \textbf{Zero-shot performance.}  As depicted in Table \ref{table:zero-shot}, all CD models pre-trained by \texttt{Ce-100K} exhibit zero-shot capability on the two benchmark datasets. This suggests that \texttt{Ce-100K} imparts generalized knowledge that can be effectively transferred within the CD domain. Furthermore, the transformer-based CD model demonstrates strong performance on the SECOND dataset, which includes a wider range of change types (30 change types). Conversely, on the SYSU-CD dataset, characterized by a simpler type of change (primarily 6 change types), Changer emerges with the most robust performance. Meanwhile, the difference in performance for all models is smaller on SYSU-CD compared to SECOND.
\par \noindent \textbf{Few-shot performance.}  As shown in Fig. \ref{fig-line-second}, the accuracies of different models are improved after pre-training using the \texttt{Ce-100K}.  Tables \ref{table:few-shot-second-1} and \ref{table:few-shot-sysucd-1} report the performances of training the CD model using only 5\%, 10\%, and 20\% samples (only-sup) of the original SECOND and SYSU-CD datasets, versus the performances after pre-training with \texttt{Ce-100K} and then fine-tuning using the same sample size as the only-sup. Note that in the only-sup experiments, for models that reused the computer vision community backbone such as Changer, BiT, and ChangeFormer, we used ImageNet's pre-training weights for initialization. Overall, \texttt{Ce-100K} shows a surprisingly few-shot performance. For the SECOND dataset, all the models show an average improvement of 24.12, 20.26, and 21.64 of F$_1$ in only-sup performance compared to those at 5\%, 10\%, and 20\% after pre-training on the \texttt{Ce-100K}. For the SYSU-CD dataset, all the models show an average improvement of 29.20, 4.58, and 3.02 of F$_1$ in only-sup performance compared to those at 5\%, 10\%, and 20\% after pre-training on the \texttt{Ce-100K}. In addition, FC-series models (FC-EF, FC-Siam-conc, and FC-Siam-diff) are affected by the category imbalance problem in CD in the few-shot case. Therefore all the metrics are 0, and the use of \texttt{Ce-100K} effectively alleviated the problem.
\par We also conducted experiments at 50\% and 100\% sample ratios (Tables \ref{table:few-shot-second-2} and \ref{table:few-shot-sysucd-2}, Fig. \ref{fig-line-second}), and the accuracy of most of the models still benefited from \texttt{Ce-100K}, and even broke through the only-sup case of 100\% samples using 50\% samples fine-tuning. This is because the \texttt{Ce-100K} could provide CD domain-specific knowledge that maximizes the pre-training gain \cite{wangpre2023}.

\section{Conclusion}\label{sec:conclusion}
\vspace{-7pt}
\par In this paper, we propose ChangeAnywhere, a novel method for generating bi-temporal change detection based on a semantic latent diffusion model and single-temporal remote sensing semantic segmentation dataset. The method is very flexible and does not need to change the training method of existing DDPMs. Moreover, the randomness of the non-change region was considered in the CD sample generation process, which makes the sample more realistic. We generated \texttt{Ce-100K}, the largest synthetic dataset for CD so far, based on ChangeAnywhere. The zero-shot and few-shot performance of the CD model pre-trained by \texttt{Ce-100K} on the two benchmark datasets proves the value of ChangeAnywhere and \texttt{Ce-100K}. The \texttt{Ce-100K} can help the SOTA CD model break through the existing accuracy. In the future, we will further validate their potential in semantic CD and develop more suitable pre-training approaches with generated samples. Meanwhile, we hope that ChangeAnywhere and \texttt{Ce-100K} can assist in the construction of remote sensing foundation models.

\section{Acknowledgements}\label{sec:acknowledgements}
\vspace{-7pt}
\par This study is supported by the National Natural Science Foundation of China Major Program (42192580).


\small
\bibliographystyle{IEEEbib}
\bibliography{refs}

\clearpage
\begin{sidewaystable}[!ht]
  \setlength\tabcolsep{5pt} \small
  \renewcommand{\arraystretch}{1.05}
  \caption{\centering{Few-shot performance evaluation of different methods on the SECOND dataset.}}
  \centering
  \begin{tabular}{l | c | l l l l | l l l l | l l l l } 
  \hline
  \multirow{2}{*}{\textbf{Method}} & \multirow{2}{*}{\textbf{Backbone}} & \multicolumn{4}{c}{\textbf{SECOND-5\%}} & \multicolumn{4}{c}{\textbf{SECOND-10\%}} & \multicolumn{4}{c}{\textbf{SECOND-20\%}}\\ 
  \cline{3-14}   
      ~ & ~ & F$_1$ & Precision & Recall & IoU & F$_1$ & Precision & Recall & IoU & F$_1$ & Precision & Recall & IoU \\ \hline\hline
      \textbf{\textit{ConvNet-based}}  & ~ & ~ & ~ & ~ & ~ & ~ & ~ & ~ & ~ & ~ & ~ & ~ & ~ \\ 
      FC-EF \cite{CayeDaudt2018} & - & - & - & - & - & - & - & - & - & - & - & - & - \\ 
      \rowcolor{gray!25}
      $+$ \texttt{Ce-100K} & ~ & 28.30  & 43.82  & 20.89  & 16.48  & 33.17  & 63.07  & 22.50  & 19.88  & 46.47  & 61.39  & 37.39  & 30.27  \\ 
      FC-Siam-Conc \cite{CayeDaudt2018} & - & - & - & - & - & - & - & - & - & - & - & - & - \\ 
      \rowcolor{gray!25}
      $+$ \texttt{Ce-100K} & ~ & 16.87  & 32.76  & 11.36  & 9.21  & 39.29  & 56.51  & 30.12  & 24.45  & 46.57  & 63.04  & 36.92  & 30.35  \\ 
      FC-Siam-Diff \cite{CayeDaudt2018} & - & - & 74.07  & - & - & - & - & - & - & - & - & - & - \\ 
      \rowcolor{gray!25}
      $+$ \texttt{Ce-100K} & ~ & 29.25  & 61.76  & 19.16  & 17.13  & 29.14  & 71.53  & 18.30  & 17.06  & 31.96  & 77.09  & 20.16  & 19.02  \\ 
      SNUNet \cite{Fang2022SNUNet} & - & 47.99  & 58.91  & 40.48  & 31.57  & 55.14  & 56.97  & 53.41  & 38.06  & 58.95  & 60.39  & 57.57  & 41.79  \\ 
      \rowcolor{gray!25}
      $+$ \texttt{Ce-100K} & ~ & 57.91  & 64.62  & 52.46  & 40.75  & 60.62  & 64.38  & 57.26  & 43.49  & 63.16  & 66.68  & 59.99  & 46.15  \\ 
      LightCDNet \cite{10214556} & - & 27.70  & 59.20  & 18.08  & 16.08  & 41.27  & 57.62  & 32.15  & 26.00  & 42.65  & 58.56  & 33.54  & 27.11  \\ 
      \rowcolor{gray!25}
      $+$ \texttt{Ce-100K} & ~ & 60.25  & 65.32  & 55.92  & 43.12  & 62.88  & 64.12  & 61.68  & 45.85  & 63.87  & 67.98  & 60.22  & 46.91  \\
      Changer \cite{fang2022changer} & ResNet18 & 9.73  & 37.68  & 5.59  & 5.11  & 47.01  & 61.83  & 37.93  & 30.73  & 53.52  & 64.92  & 45.52  & 36.54  \\ 
      \rowcolor{gray!25}
      $+$ \texttt{Ce-100K} & ~ & 62.22  & 66.57  & 58.41  & 45.16  & 64.77  & 66.28  & 63.33  & 47.90  & 67.03  & 69.85  & 64.42  & 50.41  \\ 
      \textbf{\textit{Transformer-based}} & ~ & ~ & ~ & ~ & ~ & ~ & ~ & ~ & ~ & ~ & ~ & ~ & ~ \\
      BiT \cite{Chen2022Remote} & ResNet18 & 44.84  & 56.46  & 37.18 & 28.90  & 52.15  & 59.75  & 46.26  & 35.27  & 58.77  & 61.71  & 56.09  & 41.61  \\ 
      \rowcolor{gray!25}
      $+$ \texttt{Ce-100K} & ~ & 62.65  & 70.51  & 56.37  & 45.62  & 65.06  & 69.30  & 61.30  & 48.21  & 66.15  & 73.11  & 60.40  & 49.42  \\ 
      ChangeFormer \cite{bandara2022transformerbased} & MiT-B1 & 60.35  & 71.91  & 51.99  & 43.21  & 65.00  & 70.81  & 60.07  & 48.15  & 66.71  & 73.77  & 60.88  & 50.05  \\ 
      \rowcolor{gray!25}
      $+$ \texttt{Ce-100K} & ~ & 66.13  & 69.54  & 63.05  & 49.40  & 67.72  & 71.78  & 64.09  & 51.19  & 68.54  & 72.16  & 65.26  & 52.13  \\ 
      \hline
    \end{tabular}
    \label{table:few-shot-second-1}
\end{sidewaystable}

\begin{sidewaystable}[!ht]
  \setlength\tabcolsep{5pt} \small
  \renewcommand{\arraystretch}{1.05}
  \caption{\centering{Few-shot performance evaluation of different methods on the SYSU-CD dataset.}}
  \centering
  \begin{tabular}{l | c | c c c c | c c c c | c c c c } 
  \hline
  \multirow{2}{*}{\textbf{Method}} & \multirow{2}{*}{\textbf{Backbone}} & \multicolumn{4}{c}{\textbf{SYSU-CD-5\%}} & \multicolumn{4}{c}{\textbf{SYSU-CD-10\%}} & \multicolumn{4}{c}{\textbf{SYSU-CD-20\%}}\\ 
  \cline{3-14}   
      ~ & ~ & F$_1$ & Precision & Recall & IoU & F$_1$ & Precision & Recall & IoU & F$_1$ & Precision & Recall & IoU \\ \hline\hline
      \textbf{\textit{ConvNet-based}}  & ~ & ~ & ~ & ~ & ~ & ~ & ~ & ~ & ~ & ~ & ~ & ~ & ~ \\ 
      FC-EF \cite{CayeDaudt2018} & - & - & - & - & - & 66.17 & 74.41 & 59.57 & 49.44 & 70.04 & 70.74 & 69.36 & 53.9 \\ 
      \rowcolor{gray!25}
      $+$ \texttt{Ce-100K} & ~ & 74.40  & 73.26  & 75.58  & 59.24  & 75.03  & 73.85  & 76.24  & 60.03  & 75.36  & 76.71  & 74.05  & 60.46  \\ 
      FC-Siam-Conc \cite{CayeDaudt2018} & - & - & - & - & - & 68.61 & 72.95 & 64.76 & 52.22 & 71.65 & 74.57 & 68.95 & 55.83 \\
      \rowcolor{gray!25}
      $+$ \texttt{Ce-100K} & ~ & 67.36  & 76.86  & 59.95  & 50.78  & 70.36  & 76.19  & 65.36  & 54.28  & 71.50  & 78.06  & 65.96  & 55.64  \\
      FC-Siam-Diff \cite{CayeDaudt2018} & - & - & - & - & - & 59.86 & 78.81 & 48.26 & 42.72 & 58.84 & 82.02 & 45.87 & 41.68 \\
      \rowcolor{gray!25}
      $+$ \texttt{Ce-100K} & ~ & 55.56  & 80.68  & 42.37  & 38.47  & 57.25  & 77.93  & 45.24  & 40.10  & 56.17  & 85.27  & 41.88  & 39.05  \\
      SNUNet \cite{Fang2022SNUNet} & - & 69.65  & 74.86  & 65.12  & 53.44  & 73.70  & 77.81  & 70.01  & 58.36  & 75.40  & 78.02  & 72.95  & 60.51  \\
      \rowcolor{gray!25}
      $+$ \texttt{Ce-100K} & ~ & 76.40  & 80.85  & 72.41  & 61.81  & 78.28  & 81.20  & 75.56  & 64.31  & 78.90  & 82.31  & 75.76  & 65.16  \\
      LightCDNet \cite{10214556}  & - & 66.67  & 79.07  & 57.64  & 50.01  & 71.55  & 77.71  & 66.29  & 55.70  & 72.45  & 81.97  & 64.91  & 56.80  \\
      \rowcolor{gray!25}
      $+$ \texttt{Ce-100K} & ~ & 79.80  & 79.02  & 80.60  & 66.39  & 80.82  & 79.75  & 81.93  & 67.82  & 80.19  & 82.17  & 78.31  & 66.94  \\
      Changer \cite{fang2022changer} & ResNet18 & 66.59  & 78.29  & 57.93  & 49.91  & 72.06  & 80.07  & 65.50  & 56.32  & 73.85  & 79.95  & 68.62  & 58.55  \\
      \rowcolor{gray!25}
      $+$ \texttt{Ce-100K} & ~ & 79.65  & 81.93  & 77.48  & 66.18  & 80.67  & 83.55  & 77.97  & 67.60  & 81.72  & 84.43  & 79.45  & 69.10  \\
      \textbf{\textit{Transformer-based}} & ~ & ~ & ~ & ~ & ~ & ~ & ~ & ~ & ~ & ~ & ~ & ~ & ~ \\ 
      BiT \cite{Chen2022Remote} & ResNet18 & 69.65  & 69.12  & 70.19  & 53.44  & 69.41  & 67.90  & 70.98  & 53.15  & 74.48  & 74.09  & 74.88  & 59.34  \\ 
      \rowcolor{gray!25}
      $+$ \texttt{Ce-100K} & ~ & 73.59  & 78.23  & 69.46  & 58.21  & 75.12  & 81.71  & 69.52  & 60.16  & 77.48  & 81.44  & 73.89  & 63.24  \\ 
      ChangeFormer \cite{bandara2022transformerbased} & MiT-B1 & 78.35  & 80.60  & 76.23  & 64.41  & 78.68  & 86.57  & 72.11  & 64.86  & 80.92  & 87.04  & 75.60  & 67.95  \\ 
      \rowcolor{gray!25}
      $+$ \texttt{Ce-100K} & ~ & 77.74  & 86.09  & 70.87  & 63.59  & 79.13  & 87.30  & 72.36  & 65.47  & 80.50  & 87.82  & 74.31  & 67.37  \\ 
      \hline
  \end{tabular}
  \label{table:few-shot-sysucd-1}
\end{sidewaystable}

\begin{sidewaystable}[!ht]
  \setlength\tabcolsep{5pt} \small
  \renewcommand{\arraystretch}{1.05}
  \caption{\centering{Performance evaluation of different methods on the SECOND dataset.}}
  \centering
  \begin{tabular}{l | c | c c c c | c c c c  }
  \hline
  \multirow{2}{*}{\textbf{Method}} & \multirow{2}{*}{\textbf{Backbone}} & \multicolumn{4}{c}{\textbf{SECOND-50\%}} & \multicolumn{4}{c}{\textbf{SECOND-100\%}}\\ 
  \cline{3-10} 
      ~ & ~ & F$_1$ & Precision & Recall & IoU & F$_1$ & Precision & Recall & IoU \\ \hline \hline
      \textbf{\textit{ConvNet-based}}  & ~ & ~ & ~ & ~ & ~ & ~ & ~ & ~ & ~ \\
      FC-EF \cite{CayeDaudt2018} & - & - & - & - & - & 47.58  & 61.58  & 38.77  & 31.22 \\
      \rowcolor{gray!25}
      $+$ \texttt{Ce-100K} & ~ & 52.46  & 59.64  & 46.83  & 35.56  & 54.57  & 62.07  & 48.68  & 37.52  \\
      FC-Siam-Conc \cite{CayeDaudt2018} & - & 44.38  & 58.69  & 35.69  & 28.52  & 53.16  & 60.85  & 47.19  & 36.20  \\
      \rowcolor{gray!25}
      $+$ \texttt{Ce-100K} & ~ & 54.78  & 63.96  & 47.91  & 37.73  & 56.71  & 65.74  & 49.86  & 39.58  \\ 
      FC-Siam-Diff \cite{CayeDaudt2018} & - & 35.18  & 68.09  & 23.71  & 21.34  & 46.40  & 69.19  & 34.91  & 30.21  \\
      \rowcolor{gray!25}
      $+$ \texttt{Ce-100K} & ~ & 41.99  & 77.38  & 28.82  & 26.58  & 47.63  & 77.99  & 34.28  & 31.26  \\ 
      SNUNet \cite{Fang2022SNUNet} & - & 62.51  & 64.32  & 60.80  & 45.47  & 65.69  & 68.02  & 63.51  & 48.91  \\ 
      \rowcolor{gray!25}
      $+$ \texttt{Ce-100K} & ~ & 65.92  & 69.17  & 62.97  & 49.17  & 67.31  & 71.04  & 63.96  & 50.73  \\ 
      LightCDNet \cite{10214556}  & - & 47.62  & 61.19  & 38.98  & 31.25  & 52.03  & 62.80  & 44.42  & 35.17  \\ 
      \rowcolor{gray!25}
      $+$ \texttt{Ce-100K} & ~ & 66.58  & 70.46  & 63.10  & 49.90  & 68.50  & 71.40  & 65.82  & 52.09  \\ 
      Changer \cite{fang2022changer} & ResNet18 & 61.31  & 66.44  & 96.91  & 44.20  & 64.16  & 69.29  & 59.73  & 47.23  \\ 
      \rowcolor{gray!25}
      $+$ \texttt{Ce-100K} & ~ & 69.53  & 72.27  & 67.00  & 53.30  & 70.31  & 73.39  & 67.48  & 54.22  \\ 
      \textbf{\textit{Transformer-based}} & ~ & ~ & ~ & ~ & ~ & ~ & ~ & ~ & ~ \\ 
      BiT \cite{Chen2022Remote} & ResNet18 & 63.49 & 70.31 & 57.87 & 46.51 & 66.29  & 73.40  & 60.43  & 49.58  \\ 
      \rowcolor{gray!25}
      $+$ \texttt{Ce-100K} & ~ & 67.41  & 75.22  & 61.07  & 50.84  & 68.75  & 75.72  & 62.96  & 52.39  \\ 
      ChangeFormer \cite{bandara2022transformerbased} & MiT-B1 & 70.51  & 74.24  & 67.13  & 54.45  & 70.26  & 73.44  & 67.34  & 54.15  \\ 
      \rowcolor{gray!25}
      $+$ \texttt{Ce-100K} & ~ & 71.32  & 75.65  & 67.46  & 55.42  & 71.19  & 74.43  & 68.22  & 55.26  \\
      \hline
  \end{tabular}
  \label{table:few-shot-second-2}
\end{sidewaystable}

\begin{sidewaystable}[!ht]
  \setlength\tabcolsep{5pt} \small
  \renewcommand{\arraystretch}{1.05}
  \caption{\centering{Performance evaluation of different methods on the SYSU-CD dataset.}}
  \centering
  \begin{tabular}{l | c | c c c c | c c c c  }
  \hline
  \multirow{2}{*}{\textbf{Method}} & \multirow{2}{*}{\textbf{Backbone}} & \multicolumn{4}{c}{\textbf{SYSU-CD-50\%}} & \multicolumn{4}{c}{\textbf{SYSU-CD-100\%}}\\ 
  \cline{3-10} 
      ~ & ~ & F$_1$ & Precision & Recall & IoU & F$_1$ & Precision & Recall & IoU \\ \hline \hline
      \textbf{\textit{ConvNet-based}}  & ~ & ~ & ~ & ~ & ~ & ~ & ~ & ~ & ~ \\
      FC-EF \cite{CayeDaudt2018} & - & 71.47 & 74.73 & 68.48 & 55.6 & 73.99  & 78.50  & 69.98  & 58.72  \\ 
      \rowcolor{gray!25}
      $+$ \texttt{Ce-100K} & ~ & 76.64  & 75.21  & 78.13  & 62.13  & 77.42  & 76.70  & 78.16  & 63.16  \\ 
      FC-Siam-Conc \cite{CayeDaudt2018} & - & 73.51 & 78.43 & 69.18 & 58.12 & 75.62  & 78.37  & 73.06  & 60.80  \\ 
      \rowcolor{gray!25}
      $+$ \texttt{Ce-100K} & ~ & 73.15  & 78.77  & 68.28  & 57.67  & 73.68  & 80.75  & 67.75  & 58.33  \\ 
      FC-Siam-Diff \cite{CayeDaudt2018} & - & 60.38 & 87.76 & 46.02 & 43.25 & 60.05  & 89.95  & 45.06  & 42.90  \\
      \rowcolor{gray!25}
      $+$ \texttt{Ce-100K} & ~ & 65.07  & 86.81  & 52.04  & 48.23  & 65.89  & 88.21  & 52.59  & 49.14  \\
      SNUNet \cite{Fang2022SNUNet} & - & 78.75  & 82.91  & 74.98  & 64.94  & 80.05  & 85.16  & 75.51  & 66.73  \\ 
      \rowcolor{gray!25}
      $+$ \texttt{Ce-100K} & ~ & 80.67  & 83.28  & 78.22  & 67.60  & 80.33  & 81.22  & 79.46  & 67.12  \\
      LightCDNet \cite{10214556}  & - & 77.50  & 79.28  & 75.80  & 63.27  & 77.14  & 80.12  & 74.37  & 62.78  \\ 
      \rowcolor{gray!25}
      $+$ \texttt{Ce-100K} & ~ & 82.03  & 83.48  & 80.62  & 69.53  & 82.42  & 83.01  & 81.83  & 70.09  \\
      Changer \cite{fang2022changer} & ResNet18 & 77.60  & 81.95  & 73.69  & 63.40  & 79.47  & 80.19  & 78.77  & 65.94  \\ 
      \rowcolor{gray!25}
      $+$ \texttt{Ce-100K} & ~ & 83.25  & 85.87  & 80.78  & 71.30  & 83.52  & 84.91  & 82.17  & 71.70  \\ 
      \textbf{\textit{Transformer-based}} & ~ & ~ & ~ & ~ & ~ & ~ & ~ & ~ & ~ \\ 
      BiT \cite{Chen2022Remote} & ResNet18 & 72.97 & 65.67 & 82.09 & 57.44 & 74.83  & 73.92  & 75.76  & 59.78  \\ 
      \rowcolor{gray!25}
      $+$ \texttt{Ce-100K} & ~ & 77.22  & 83.29  & 71.98  & 62.89  & 78.85  & 78.74  & 78.96  & 65.08  \\ 
      ChangeFormer \cite{bandara2022transformerbased} & MiT-B1 & 81.41  & 88.72  & 75.21  & 68.65  & 81.20  & 88.93  & 74.71  & 68.35  \\
      \rowcolor{gray!25}
      $+$ \texttt{Ce-100K} & ~ & 82.34  & 89.00  & 76.61  & 69.98  & 81.75  & 88.01  & 76.32  & 69.13  \\ 
      \hline
  \end{tabular}
  \label{table:few-shot-sysucd-2}
\end{sidewaystable}

\end{document}